\documentclass[conference]{IEEEtran}
\usepackage{cite}
\usepackage{amsmath,amssymb,amsfonts}
\usepackage{algorithmic}
\usepackage{graphicx}
\usepackage{textcomp}
\usepackage{xcolor}
\usepackage{multirow}
\usepackage{graphicx}
\usepackage{todonotes}

\def\BibTeX{{\rm B\kern-.05em{\sc i\kern-.025em b}\kern-.08em
    T\kern-.1667em\lower.7ex\hbox{E}\kern-.125emX}}
\begin{document}

\title{Graph Coordinates and Conventional Neural Networks - An  Alternative for Graph Neural Networks\\

}

\author{\IEEEauthorblockN{Zheyi Qin}
\IEEEauthorblockA{\textit{Electrical and Computer Engineering} \\
\textit{Colorado State University}\\
Fort Collins, USA \\
zheyi.qin@colostate.edu}
\and
\IEEEauthorblockN{Randy Paffenroth}
\IEEEauthorblockA{\textit{Department of Data Science Program} \\
\textit{Worcester Polytechnic Institute}\\
Worcester, MA, USA \\
rcpaffenroth@wpi.edu}
\and
\IEEEauthorblockN{Anura P. Jayasumana}
\IEEEauthorblockA{\textit{Electrical and Computer Engineering} \\
\textit{Colorado State University}\\
Fort Collins, USA \\
anura.jayasumana@colostate.edu}
}

\maketitle

\begin{abstract}
Graph-based data present unique challenges and opportunities for machine learning. Graph Neural Networks (GNNs), and especially those algorithms that capture graph topology through message passing for neighborhood aggregation, have been a leading solution. However, these networks often require substantial computational resources and may not optimally leverage the information contained in the graph's topology, particularly for large-scale or complex graphs.

We propose Topology Coordinate Neural Network (TCNN) and Directional Virtual Coordinate Neural Network (DVCNN) as novel and efficient alternatives to message passing GNNs, that directly leverage the graph's topology, sidestepping the computational challenges presented by competing algorithms. Our proposed methods can be viewed as a reprise of classic techniques for graph embedding for neural network feature engineering, but they are novel in that our embedding techniques leverage ideas in Graph Coordinates (GC) that are lacking in current practice. 

Experimental results, benchmarked against the Open Graph Benchmark Leaderboard, demonstrate that TCNN and DVCNN achieve competitive or superior performance to message passing GNNs. For similar levels of accuracy and ROC-AUC, TCNN and DVCNN need far fewer trainable parameters than contenders of the OGBN Leaderboard. The proposed TCNN architecture requires fewer parameters than any neural network method currently listed in the OGBN Leaderboard for both OGBN-Proteins and OGBN-Products datasets. Conversely, our methods achieve higher performance for a similar number of trainable parameters. These results hold across diverse datasets and edge features, underscoring the robustness and generalizability of our methods. By providing an efficient and effective alternative to message passing GNNs, our work expands the toolbox of techniques for graph-based machine learning. A significantly lower number of tunable parameters for a given evaluation metric makes TCNN and DVCNN especially attractive for resource limited IoT/mobile devices and for reducing power consumption of ML models.  

\end{abstract}

\section{Introduction}\label{Introduction}

Node and link property predictions are a fundamental problem in graph-based machine learning and network analysis, aiming to predict the attributes or properties of nodes in a given graph. Graphs are mathematical structures consisting of nodes (or vertices) and edges, and are highly expressive representations of complex data and relationships. They have found immense utility in a wide range of domains, such as social networks, biological systems, and computer vision, to name a few. Recent years have seen significant progress in the development of graph-based machine learning techniques, including message passing based Graph Neural Networks (GNNs)\cite{zhou2020graph}, graph embedding based Neural Networks \cite{grover2016node2vec}, Graph Convolutional Networks (GCNs)\cite{kipf2016semi}, and Graph Attention Networks (GATs)\cite{velickovic2017graph}, which have proven effective in tackling the node property prediction problem. These models capture the graph's local and global structural information and the interactions between nodes and their neighbors to learn powerful representations for downstream tasks.

As observed in Wikipedia, which is at least indicative of the common parlance though perhaps not authoritative, "[T]he key design element of GNNs is the use of pairwise message passing, such that graph nodes iteratively update their representations by exchanging information with their neighbors." \footnote{https://en.wikipedia.org/wiki/Graph\_neural\_network}  Accordingly, it is important to emphasize that this paper does not derive any new methods for GNNs as defined above. However, our focus in this paper is on Graph Embedding based Neural Networks (GENN), where there is an explicit graph embedding step (e.g., into some Euclidean space) before applying 
a Neural Network, as opposed to the commonly defined GNNs.

The node property prediction task holds great promise in driving advances in various applications, including but not limited to drug discovery, fraud detection, recommender systems, and community detection. In essence, the goal is to leverage a graph's inherent structure and topological features, along with any available node or edge features, to predict node properties, which can be categorical, binary, or continuous variables.

However, there are three major limitations associated with GNNs. 
First, despite the significant achievements of GNNs in a range of graph-based tasks, their efficiency in capturing and utilizing topological information may sometimes be compromised. For instance, obtaining a comprehensive understanding of topological information within node features in large-scale or complex graphs could necessitate multiple rounds of aggregation spanning several layers. This iterative process can be computationally intensive, increasing resource consumption and extending processing time.

Second, %
these multiple layers of aggregation can lead to challenges with information propagation, particularly over long distances. It has been observed that the influence of distant nodes can get diminished during repeated aggregation and transformation operations, a phenomenon commonly referred to as the over-smoothing \cite{cai2020note} problem. This issue can potentially lead to a loss of unique node information and even to the inability of the model to distinguish between different nodes, thereby impacting the overall performance of GNNs.

The third challenge is related to their heavy resource requirements (power, memory, computation, latency).
The primary criteria for comparison of different ML approaches for a specific problem is accuracy; e.g.,  leaderboard ranking is purely based on accuracy. However, a balance between model complexity and performance is crucial in practical deployments,  such as with mobile systems or IoT \cite{MAHDAVINEJAD2018161}.    Complex ML models may not fit within the limited storage capacity of such devices. Many IoT devices are battery-powered, and ML computations can be energy-intensive. Designing ML algorithms and models that consume minimal energy is essential to extend the operational lifespan of these devices. Some IoT applications require real-time or near-real-time processing for timely decision-making.

An alternative approach for solving graph related problems via machine learning is to use a network embedding scheme to represent the graph using a low-dimensional representation, and then use this representation with a conventional neural network to solve the specific problem. However, with the advent of GNNs, this approach is treated as an 'also ran' method for ML applications related to networked data rather than an approach that is competitive to GNNs, let alone as an approach that can outperform GNNs under certain circumstances or criteria.  

For clarity, we introduce terminology that effectively segregates two classes: Graph Embedding Neural Networks (GENN) and Graph Neural Networks (GNN). GENN is characterized by an explicit embedding step, where the graph structures are converted into a fixed-size vector in a latent space before further processing. This step distinctly maps graph structures or substructures into vector representations, which are then used as inputs for subsequent neural network layers. The rationale behind GENNs is to distill the essential information of the graph into a concise representation that traditional neural architectures can readily process. In contrast, GNNs do not possess an explicit embedding step. Instead, they often pass messages among nodes to aggregate and disseminate information across the graph. The essence of GNNs lies in their ability to operate directly on the graph structure, assimilating information from neighbors and, in some cases, even distant nodes, to compute the final output. This direct operation obviates the need for an intermediary embedding, allowing GNNs to handle graphs more fluidly and adaptively.

Existing graph embedding schemes are considered to suffer computational
and statistical performance drawbacks, e.g., due to the need for eigen decomposition of large data matrices that is expensive unless the solution quality is significantly compromised with approximations, or due to their optimization objectives that are not robust to the diverse patterns observed in networks (such as homophily
and structural equivalence) \cite{grover2016node2vec}. Optimization objectives used by graph embedding techniques are addressed in \cite{cai2018comprehensive}. 
The capability of GAE frameworks to address the network embedding problem is also cited as a reason for not going through a graph embedding step \cite{wu2020comprehensive}, although this approach is also computationally expensive. 
The validity of these arguments in favor of GNNs as opposed to embedding combined with ML appears to be confirmed by many recent results. 
For example,  41 out of 76 results on OGBN Leaderboard  \cite{ogbWebsite} \cite{hu2020ogb} for OGBN-Products and OGBN-Proteins problem are based on GNNs with 25 remaining corresponding to GENN based approaches such as node2vec \cite{grover2016node2vec}.

Thus for a graph embedding scheme to be a competitive alternative, it has to be computationally efficient, and be able to capture both local and global information about graph structure with significant fidelity. This appears not to be the case with many of the well known embedding schemes \cite{grover2016node2vec}. However, this does not preclude the possibility of existence of such embedding schemes.  
In this paper, we use two specific graph coordinate based embedding schemes, Topology Coordinates (TC) and Directional Virtual Coordinates (DVC) to demonstrate that graph embedding approaches that capture the local and global graph topology effectively can in fact provide solutions that are more computationally efficient compared to GNNs that rely on message passing, while being competitive in accuracy. DVC approach is significantly different from existing embedding strategies, that it does not fit into existing embedding classifications, such as that in \cite{cai2018comprehensive}.

\subsection{Contribution}\label{Contribution}

Work presented here responds to the inherent challenges of GNNs presented earlier, including the computation requirements necessitated by the need for multiple rounds of aggregation spanning multiple layers and the challenges with information propagation over long distances,  while overcoming the limitations of existing embedding schemes which have not gained traction due to their inability to provide significant improvement in performance or computation efficiency over GNNs.
Specifically, we use two graph coordinate systems, Topology Coordinates (TC) and Directional Virtual Coordinates (DVC) that have been developed in the context of localization free operations of sensor networks including topology mapping and routing. They offer an enhanced strategy for harnessing topology information more efficiently and effectively, and  revolutionizes the perception and handling of relationships between nodes in a graph. The two embedding schemes exploit the fact that the distance matrix of a graph is low-rank, while the adjacency matrix is not. As existing literature \cite{jayasumana2019network},\cite{mahindre2020inference} demonstrate, these schemes also may be based on a relatively small set of distance samples of a graph, such as random measurements, a fact we do not exploit in this paper.

TCNN, based on a system of  Topology Coordinates %
(TC)\cite{jayasumana2019network}, and DVCNN, based on Directional Virtual Coordinates \cite{dhanapala2011directional} which revolutionizes the perception and handling of relationships between nodes in a graph and reduces the number of trainable parameters in machine learning models without compromising their performance. The primary objective of Graph Coordinate (GC, including TC and DVC) is to strike a balance between model complexity and predictive capabilities, offering a scalable and efficient solution to the ever-growing demands of modern machine learning applications. Results presented below demonstrate that these  methods are more computationally efficient than Graph Neural Networks (GNN) and Node2Vec \cite{grover2016node2vec}. %

Unlike conventional techniques, the Graph Coordinate method is rooted in the concept of strategically sampling a small yet representative fraction of relationships among nodes rather than attempting to capture the entirety of complex relationships. This alternative approach offers three %
distinct advantages. Firstly, it captures the topological essence of the graph with a much smaller, manageable set of relationships, thus reducing the computational burden associated with large-scale or intricate graphs. Secondly, this method also ensures a broader coverage of topological information, including distant relationships, overcoming the over-smoothing problem associated with the multi-layer aggregation process in GNNs. Third, GCs are easy to generate and may accurately be approximated even when complete topology information is not available\cite{mahindre2020inference}.

A remarkable aspect of the Graph Coordinate approach lies in its adaptability to non-Graph Neural Networks. Once the Topology Coordinates are extracted, they can be fed into a wide range of non-Graph Neural Network models. This opens up opportunities for utilizing the rich graph topology information more efficiently, even in machine learning models not originally designed to handle graph-structured data.

We demonstrate the TCNN and DVCNN approach's efficacy across various machine learning model architectures, emphasizing its robustness and adaptability to different structures. Our extensive experiments with benchmark datasets and comparison with results on OGBN Leaderboard  \cite{ogbWebsite} \cite{hu2020ogb} reveal that TCNN and DVCNN maintain competitive performance levels while significantly reducing the computational resources required for training, thus making it an attractive solution for large-scale and resource-constrained applications.

The remainder of the paper is organized as follows: Section \ref{related work} reviews a few relevant related works. Section \ref{Graph Coordinates Introduction} provides an introduction to our topology-coordinates-based methodology and delineates its comparison with Graph Neural Networks (GNNs). Section \ref{methodology} presents how we apply our method on single graph and multilayer graphs \cite{Wikipedia2023} for machine learning. Section \ref{Dataset and Task} presents the experimental setup, including the benchmark datasets and model architectures employed to evaluate the proposed technique. Section \ref{Efficacy of Topological Coordinates} presents that topology coordinates can efficiently capture graph topology information. Section \ref{er2} and Section \ref{er1} discuss the results of our experiments; they also highlight the benefits of TCNN and DVCNN in terms of trainable parameter reduction and performance preservation. Section \ref{section:Comparison} compares our model performance with the OGBN leaderboard. Finally, Section \ref{conclusion} concludes the paper and offers insights into potential future research directions in this domain.

\section{Related Work}\label{related work}

\subsection{Graph Neural Networks (GNNs)}

Graph Neural Networks (GNNs) represent a powerful class of neural network architectures specifically designed for handling graph-structured data. Central to the functioning of GNNs is a process known as neighborhood aggregation or message passing %
This method allows GNNs to incorporate the graph's topology, encompassing both the structure and connections, into learning informative representations for nodes and edges.

The neighborhood aggregation process in GNNs operates by permitting each node to accumulate and assimilate the features of its immediate neighbors based on the graph's topology. This aggregation procedure effectively encodes the local context of each node into its features, and by propagating these features across multiple layers, a broader global context can also be integrated into each node's representation.

In mathematical terms, the updated feature of a node $v$, denoted as $h_v'$, is computed as:

\begin{equation}
h_v' = \sigma(\mathbf{W} \cdot AGG(\{h_u : u \in \mathcal{N}(v)\}))
\end{equation}

where:

\begin{itemize}
    \item $h_u$ represents the feature of a neighboring node $u$,
    \item $\mathcal{N}(v)$ represents the set of neighbors of node $v$ according to the graph's topology,
    \item $AGG$ is an aggregation function, such as mean, sum, or max, which combines the features of the neighboring nodes,
    \item $\mathbf{W}$ is a learnable weight matrix, which transforms the aggregated feature into the new feature space,
    \item $\sigma$ is a non-linear activation function, such as ReLU or tanh.
\end{itemize}

By repeatedly applying this process across multiple layers, a GNN captures a wider context of the graph in each node's features, thereby encoding the local and global topology of the graph.

A comprehensive survey of various GNN methods, including their architectures, training mechanisms, and applications, can be found in the comprehensive survey by Wu et al. \cite{wu2020comprehensive}. Graph Echo State Network (GraphESN) \cite{gallicchio2010graph} is composed of an encoder and an output layer. The encoder, which is randomly initialized, does not require training. Instead, it employs a contractive state transition function that recurrently updates the states of the nodes until the overall state of the graph reaches convergence. Once this state is achieved, the output layer is trained by utilizing the fixed node states as inputs.

Minyi Dai et al. \cite{dai2021graph} developed a novel graph neural network (GNN) model to predict properties of polycrystalline materials. Their approach incorporated the physical features of individual grains and their interactions. The GNN model was able to achieve a low prediction error and provide insights into the importance of each feature in each grain for property prediction, showing a promising direction for using GNNs for material property prediction.

Filippo Maria Bianchi, Daniele Grattarola, and Cesare Alippi \cite{bianchi2020spectral} proposed an innovative graph clustering method that built on and addressed the limitations of spectral clustering (SC). They used a GNN to solve a continuous relaxation of the normalized minCUT problem, avoiding the need for the computationally expensive eigendecomposition of the Laplacian. Their method, which learned a clustering function, significantly improved the speed of evaluation on out-of-sample graphs and demonstrated impressive performance in both supervised and unsupervised tasks.

\subsection{Graph Embedding}
Joshua B. Tenenbaum, Vin de Silva, and John C. Langford \cite{tenenbaum2000global} present the Isomap algorithm. This method utilizes nonlinear dimensionality reduction to reveal the underlying geometric and topological characteristics of datasets with high dimensions. It achieves this by modeling them as low-dimensional manifolds embedded within high-dimensional spaces. To obtain a low-dimensional representation that preserves the intrinsic geometry of the data, Isomap follows a three-step process. Firstly, it connects data points that are close to each other in the input space, forming a neighborhood graph. Secondly, it computes the shortest paths in the neighborhood graph to estimate geodesic distances between all pairs of data points. Finally, it applies classical multidimensional scaling (MDS) to the matrix of pairwise geodesic distances.

Xiao-Dong Zhang \cite{zhang2011laplacian} offered a comprehensive survey of the eigenvalues of the Laplacian matrix of a graph. The paper discussed the Laplacian spectrum's wide range of applications, including randomized algorithms, combinatorial optimization problems, and machine learning. The survey highlighted the importance of understanding the properties of the Laplacian eigenvalues in graph analysis.

Lu Lin, Ethan Blaser, and Hongning Wang \cite{lin2022graph} explored a novel graph structural attack designed to disrupt graph spectral filters, which underpin Graph Convolutional Networks (GCNs). They introduced spectral distance, based on the eigenvalues of the graph Laplacian, to measure the disruption of spectral filters. Their attack strategy, which was effective in both black-box and white-box settings, underscored the potential vulnerabilities of GCNs and the importance of considering such attacks in the design of robust GCN models.

\subsection{Our Work}

Compared with the above works, proposed  GCNN %
employs a novel application of topology coordinates, providing a more efficient and effective way to capture graph topology information. We compare our method with the state-of-the-art GNN models in terms of performance and computational efficiency, demonstrating its robustness and the potential for scalability to larger graphs.

\section{Graph Coordinate Systems}\label{Graph Coordinates Introduction}
In this section, we outline three coordinate systems for graphs, Virtual Coordinates (VC), Topology Coordinates (TC) and Directional Virtual Coordinates (DVC). Consider a weighted graph $G$ in which $d_{jk}$ is the weighted distance between any two nodes $j$ and $k$. $d_{jk}$ is the lowest sum of weights of the edges between node $j$ and node $k$. Note that for unweighted graphs, $d_{ij}=1$ for adjacent node pairs, and is the hop distance between $i$ and $j$ for non-adjacent node pairs. Let $\mathbf{D}\in \mathbb{R}^{N\times N}$ be the matrix containing the lowest weighted distance for all two node pairs, where $N$ is the number of nodes in the graph. This distance matrix can be written as
$\mathbf{D} = [d_{ij}]$.

Virtual coordinates of a node consist of the vector of distances from a node to a set of $M$ anchor nodes. Without loss of generality, let nodes 1 to $M$ be the set of anchors, i.e., node $N_i$ and $A_i$ are synonymous for $1\leq i \leq M$. While there are anchor selection schemes in literature, for simplicity we randomly select the anchors. Thus the VC of $N_i = [ d_{{A_1}{N_i}},d_{{A_2}{N_i}} ...d_{{A_M}{N_i}}].$
The matrix $D$ can be reorganized such that its initial $M$ columns and $M$ rows correspond to the selected anchors while maintaining its diagonal elements as zeros. This leads to the following block matrix representation of

\begin{equation*}
	\mathbf{D} = \begin{bmatrix} \mathbf{A} &\quad \mathbf{B}^{T} \\
		\mathbf{B} &\quad \mathbf{C} \end{bmatrix},
\end{equation*}

where the sub-matrix $\mathbf{A} \in \mathbb{R}^{M \times M}$ consists of the weighted distances between the $M$ anchors themselves. The sub-matrix $\mathbf{B} \in \mathbb{R}^{(N-M) \times M}$ includes the weighted distances between the $M$ anchors and the remaining $N-M$ non-anchor nodes. Lastly, the sub-matrix $\mathbf{C} \in \mathbb{R}^{(N-M) \times (N-M)}$ encompasses the weighted distances among the $N-M$ non-anchor nodes themselves. Let $\mathbf{P}$ denote the $N\times M$  VC matrix. The partial distance matrix $\mathbf{P}$ can be written as 

\begin{equation*}
	\mathbf{P} = \begin{bmatrix} \mathbf{A}\\
		\mathbf{B} \end{bmatrix},
\end{equation*}
where $j$-th row vector represents the shortest weighted distance from node $j$ to all selected anchors, and is commonly known as the Virtual Coordinates of node $j$. Because we only use partial distance matrix $P$ for later steps, we only need to sample the weighted distance to anchors, not all the nodes, which is easy to sample. In the context of networking, they are obtained by broadcast messages originating at anchor nodes, whereas in the case of graph analytics, they may be obtained via the Dijkstra algorithm. It is well known that the distance matrix of a graph is low rank, whereas the adjacency matrix, which is the basis of message passing in a GNN is high rank \cite{jayasumana2019network}. Therefore,  a fraction of columns of $D$ can capture the entire topology information \cite{mahindre2020inference}. Also note that notions such as metric dimension and link dimension of a graph and the corresponding anchor sets, the resolution set and the resolving set respectively,  that allow the characterization of nodes and edges uniquely, and thus able to capture aspects of graph topology,  are directly related to VCs.

\subsection{Toplogy Coordinate System (TCS)} \label{toplogy coordinates}
Topology coordinates of a graph are obtained by 
the Singular Value Decomposition (SVD) of $P$ \cite{dhanapala2010},\cite{dhanapala2013},  

\[
\mathbf{P} = \mathbf{U} \mathbf{\Sigma} \mathbf{V}^T
\]

where,
\begin{itemize}
	\item $\mathbf{U} \in \mathbb{R}^{N \times M}$ is a unitary matrix whose columns are the eigenvectors of $\mathbf{P}\mathbf{P}^T$, $N$ is the number of nodes,
	\item $\mathbf{\Sigma} \in \mathbb{R}^{M \times M}$ is a diagonal matrix containing the square roots of eigenvalues from $\mathbf{U}$ or $\mathbf{V}$ along the diagonal. The values on the diagonal are known as the singular values of the original matrix $P$,
	\item $\mathbf{V}^T \in \mathbb{R}^{M \times M}$ (the transpose of $\mathbf{V}$) is a unitary matrix whose columns are the eigenvectors of $\mathbf{P}^T \mathbf{P}$.
\end{itemize}

The submatrix consisting of the first $n_{c}$ columns of $\mathbf{U}\mathbf{\Sigma}$, i.e., the most significant $n_{c}$ principal components,  is the topology coordinate (TC) matrix  for the graph \cite{dhanapala2013}. 
The TC of node $N_d$ is given by
\begin{equation}
    C_T(N_d) = [\mathbf{U}\mathbf{\Sigma}_{N_d,1}, \mathbf{U}\mathbf{\Sigma}_{N_d,2},..., \mathbf{U}\mathbf{\Sigma}_{N_d,n_c}]
\end{equation}
where $\mathbf{U}\mathbf{\Sigma}_{N_d,j}$ is the $N_d$-th row, $j$-th column element in $\mathbf{U}\mathbf{\mathbf{\Sigma}}$.
Past studies as well as results presented below show that a good embedding can be obtained with  $n_{c}<< N$. Therefore one may select a value of 1-2\% of $N$. However, to further reduce the value of $n_{c}$. it may also be chosen based on how much variance the topology coordinates capture. Let $\sigma_{j}$ be the $j^{th}$ singular value and $p$ be the variance of TCs we want to capture. Then, %
$n_{c}$ may be chosen such that $\sum_{j=1}^{j=n_{c}} \sigma_{j}^2 > p \times (\sum_{j=1}^{j=M} \sigma_{j}^2)$. %
\begin{equation}
    C_T(N_d) = [\mathbf{U}\mathbf{\Sigma}_{N_d,1}, \mathbf{U}\mathbf{\Sigma}_{N_d,2},..., \mathbf{U}\mathbf{\Sigma}_{N_d,n_c}]
\end{equation}
where $\mathbf{U}\mathbf{\Sigma}_{N_d,j}$ is the $N_d$-th row, $j$-th column element in $\mathbf{U}\mathbf{\mathbf{\Sigma}}$.

TC provides an embedding that captures the local as well as the global information of a node with respect to the topology of the graph. TCs, originally developed for operations such as topology mapping, routing and distance free localization in sensor networks \cite{dhanapala2013,dhanapala2010,jayasumana2019network}, have later been used in different contexts such as for sampling and compressed representation of social networks \cite{mahindre2020inference}. It has many similarities to Isomaps that were developed in the context of identifying lower dimension manifolds, whereas the TCs were developed in the context of recovering physical layouts from the communication topology of networks.

\subsection{Directional Virtual Coordinate System (DVCS)}\label{introDVCS}
DVCS is a second coordinate system that is derived from VCs for network functions\cite{dhanapala2011directional}. Its main advantage is the simplicity,  notably, DVCS obviates the necessity of Singular Value Decomposition (SVD). Each ordinate in DVCS  defines a virtual direction in the graph based on distances from two anchors, which are randomly selected in our work. Suppose the $k^{th}$ ordinate is based on the two anchors $A_p$ and $A_q$.
Then the corresponding ordinate for node $N_d$ is given by \cite{dhanapala2011directional}
\begin{equation}
{{l_k}(N_d)}={\frac{1}{2d_{A_{p}A_{q}}}}(d_{N_{d}A_{p}}-d_{N_{d}A_{q}})(d_{N_{d}A_{p}}+d_{N_{d}A_{q}})
\end{equation}
In this equation, the term $(d_{N_{d}A_{p}}-d_{N_{d}A_{q}})(d_{N_{d}A_{p}}+d_{N_{d}A_{q}})$ 
yields the position information along the virtual direction $A_p$ and $A_q$, while $({1}/{2d_{A_{p}A_{q}}})$ serves as a normalizing factor, calibrating the distance so that there is a unit difference in the ordinate between adjacent nodes.
Thus each node $N_d$ is defined by a directional virtual coordinate vector
\begin{equation}
    C_D(N_d) = [{{{l_1}(N_d)}},{{l_2}(N_d)}, ..., {{l_{n_c}}(N_d)}]
\end{equation}
The remarkable advantage of  DVCS  is its avoidance of the SVD, which reduces computational overhead. Furthermore, DVCS is efficient, requiring only a small number of anchor points to compute the partial distance matrix, thereby lowering the overall computational cost. Note that $M$ anchors yield $^{M}C_{2}$ anchor pairs. Methods for dimensionality reduction by proper selection of anchor pairs can be found in \cite{dhanapala2011directional}.

\section{Machine Learning with Graph Coordinate based Embedding}\label{methodology}
As the datasets outlined in Section \ref{Dataset and Task} indicate, often the same set of nodes are associated with multiple graph topologies. Therefore we first present the method for single graphs and later extend it to multiple graphs.
\subsection{Single Graph}
\begin{figure}[htbp]
    \centerline{\includegraphics[width=0.5\textwidth]{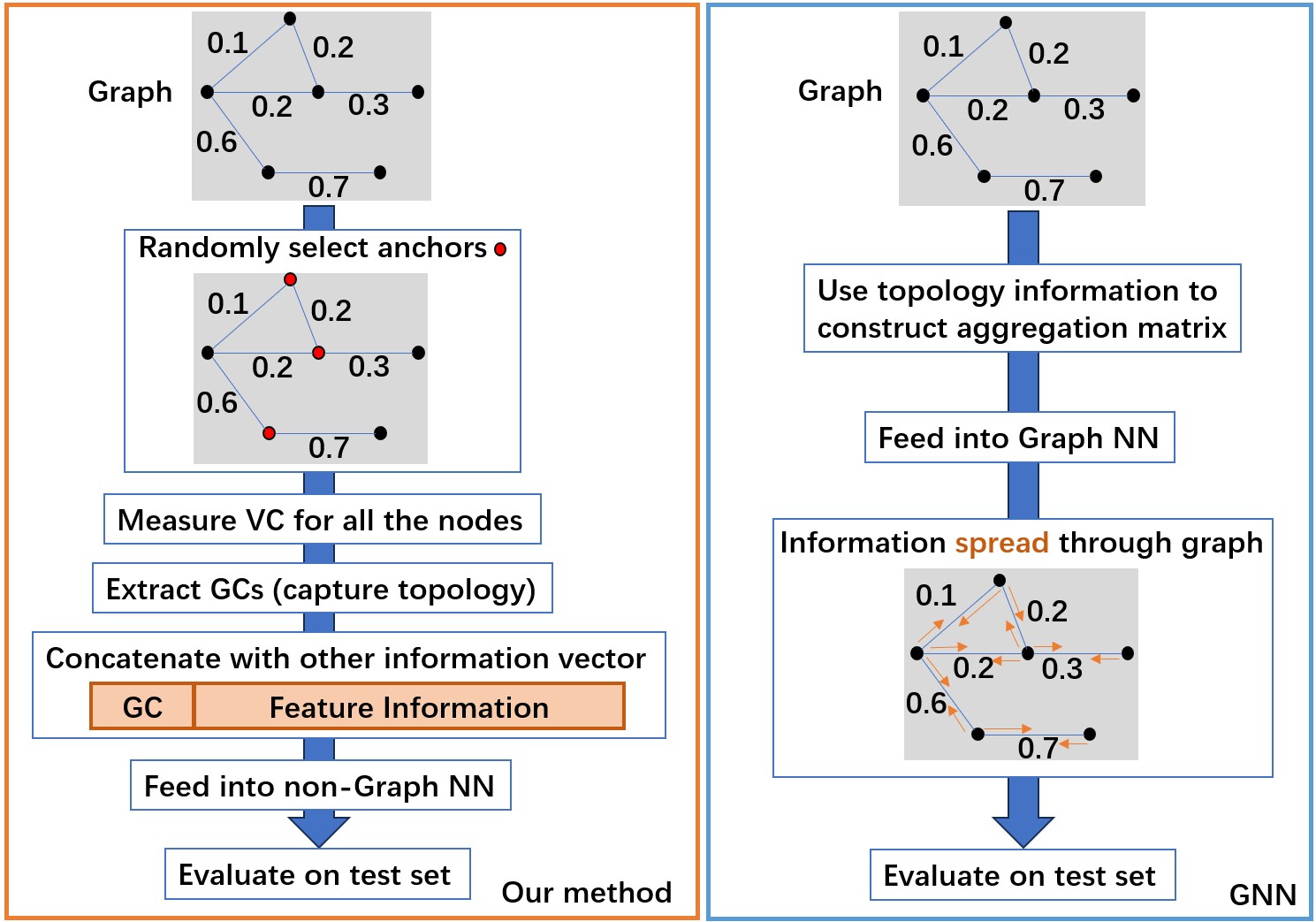}}
    \caption{TCS (left) steps compared to GNN (right)}
    \label{fig:TCScompareGNN}
\end{figure}

Drawing a contrast to Graph Neural Networks (GNN), our methodology, as depicted in Fig. \ref{fig:TCScompareGNN}, engages distinct graph coordinate (GC) systems to %
incorporate topology information. Instead of using the %
node adjacency information to propagate and aggregate node features during the learning process of  GNNs, our method extracts topology information in the form of graph coordinates (  $C_T(N_d)$ for TCS or $C_D(N_d)$ for DVCS) to feed a  neural network model.  This fundamental difference in our approach facilitates efficient utilization of topology information while significantly reducing the computational overhead, rendering our method to be both effective and much more computationally efficient. We use $C_G(N_d)$  below to denote the two graph coordinate schemes, where $G$ stands for $T$ or $D$ as appropriate, 

Fig. \ref{fig:TCScompareGNN} (left) illustrates how we concatenate graph coordinates ($C_G$) %
to get a representation vector for each node. If additional node feature vectors $NF$ are present, we concatenate them 
with the coordinates to form the vector $[C_G(N_d), NF]$, the normalized form of which serves as our input into a conventional neural network for further processing. %
Normalization of constituent element $x$ is carried out using    $  \hat{x} = [({x - \mu_{\text{train}}})/{\sigma_{\text{train}}}]$ 
where the mean value $\mu_{\text{train}}$ and the standard deviation $\sigma_{\text{train}}$ computed from the training set.
Such a normalization scheme helps to standardize the input data, improving the overall stability and performance of the neural network model.

Then we apply a neural network model with normal linear layers, where $\hat{x}$ as model input, $h_i$ as each hidden layer output, and $y$ as model output. $\mathbf{W}_{i}$ and $b_{i}$ are the weight matrix and bias vector for the i-th layer, respectively.
\begin{align*}
h_{1} &= \text{ELU}(\mathbf{W}_{1}\hat{x} + b_{1}) \\
h_{2} &= \text{ELU}(\mathbf{W}_{2}h_{1} + b_{2}) \\
&\vdots \\
h_{T-1} &= \text{ELU}(\mathbf{W}_{T-1}h_{T-2} + b_{T-1})\\
y &= \text{ELU}(\mathbf{W}_{T}h_{T-1} + b_{T})
\end{align*}
where Exponential Linear Unit (ELU) activation function is 
\begin{align*}
\text{ELU}(x) = 
\begin{cases} 
x & \text{if } x > 0 \\
\alpha(e^x - 1) & \text{if } x \leq 0 
\end{cases}
\end{align*}
Then we apply sigmoid or softmax to $y$ based on the purpose of each dataset.

\subsection{Multilayer Graphs} 
\begin{figure}[htbp]
    \centerline{\includegraphics[width=0.5\textwidth]{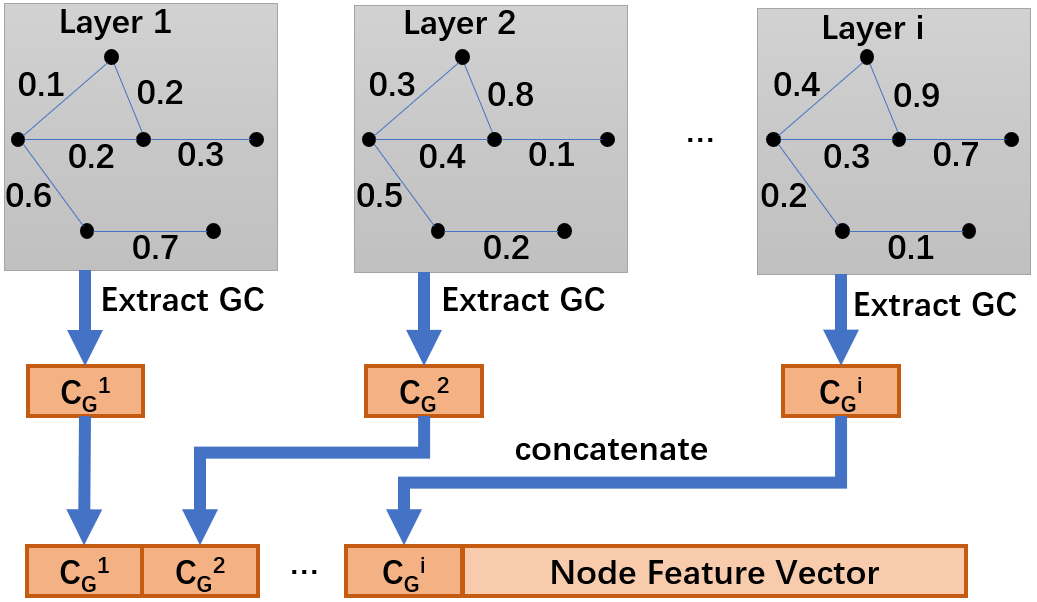}}
    \caption{Graph coordinates (GC) extraction and representation vectors concatenation for  multilayer graphs}
    \label{catVec_multigraph}
\end{figure}

For many graph problems,  such as OGBN proteins, are associated with multilayer graphs \cite{Wikipedia2023}, where each edge is associated with a node pair as well as layers (or dimensions) corresponding to different features. Such a graph can be considered as consisting of one topology in each layer, with the nodes being common. 
We denote the graph coordinates %
for the i-th feature by $C_G^i()$. 
Fig \ref{catVec_multigraph} illustrates how these different coordinates and their feature values are concatenated thus capturing node information for multiple graphs in the same vector  $[C_G^{1}(N_d),...C_G^{i}(N_d), NF]$.

\section{Datasets and Tasks}\label{Dataset and Task}
We utilize two datasets for our experiments aimed at evaluating TCNN and DVCNN. The first, OGBN-product dataset \cite{Bhatia16,chiang2019cluster}, represents an Amazon product co-purchasing network, where nodes correspond to products sold and edges indicate that they are frequently purchased together. This corresponds to a simple graph topology, i.e., one with undirected and unweighted edges, with 2,449,029 nodes and 61,859,140 edges. The node features have been generated by processing product descriptions using a bag-of-words approach, followed by Principal Component Analysis (PCA) to reduce the dimensionality of the features %
resulting in a 100-dimensional feature vector for each node, effectively capturing relevant information for the prediction task. 
The prediction task involves determining a product's category in a multi-class classification setup, utilizing the 47 top-level categories as target labels.

The second dataset, the OGBN-proteins dataset \cite{szklarczyk2019string,gene2019gene}, comprises a graph with 132,534 nodes and 39,561,252 edges, representing an undirected, weighted, and typed (according to species) graph. In this graph, nodes correspond to proteins, while edges signify various biologically meaningful associations between proteins, such as physical interactions, co-expression, or homology. Each edge is characterized by an 8-dimensional feature vector, where each dimension indicates the approximate confidence level of a single association type, with values ranging from 0 to 1. We treat it as a multilayer graph with 8 layers. 
The prediction task involves determining the presence of protein functions in a multi-label binary classification setup. In total, there are 112 distinct labels to predict. The evaluation metric employed is the average of ROC-AUC scores across all 112 tasks, which provides a comprehensive measure of the model's performance.

To assess the performance of our model across different species, and to provide a fair comparison with existing approaches, we use the same dataset splitting as indicated on OGBN websites \cite{OGBNodeProp} to split the protein nodes into training, validation, and test sets according to the species from which the proteins originate. This dataset-splitting strategy ensures a more rigorous evaluation of the model's capabilities in cross-species generalization.

\subsection{Dataset Configuration for ML}

\subsubsection{OGBN-Product}\label{dataprepro_product}
OGBN-products is an undirected graph and an unweighted graph, and this graph has 52658 connected components. We extract graph coordinates from the largest connected component and in this work, we set virtual coordinates for the nodes outside the largest component as all 0. As the largest component includes 2385902 nodes (97.42\% nodes), approximately 2.5\% of the nodes are mapped to the origin. %
We note that there is significant room for improvement here, e.g., by placing different connected components at different locations in TC space.%

\begin{figure}[htbp]
    \centerline{\includegraphics[width=0.30\textwidth]{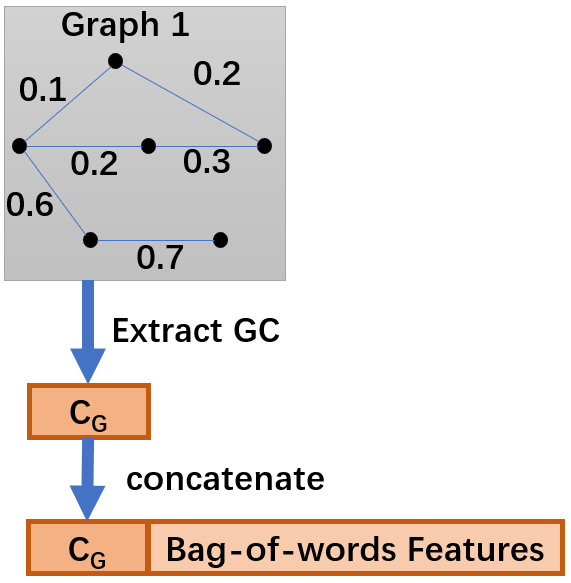}}
    \caption{Graph coordinates (GC) extraction and representation vectors concatenation for OGBN-Product dataset}
    \label{catVec_product}
\end{figure}

We use the method described in section \ref{Graph Coordinates Introduction} and \ref{methodology} for topology coordinates extraction and concatenation. Fig. \ref{catVec_product} illustrates how we concatenate graph coordinates (GC) (topology coordinates (TC) or directional virtual coordinates (DVC)) 
to get a representation vector for each node in the OGBN-Product graph. We treat the graph data as a single graph and concatenate the extracted graph coordinates (TC or DVC) with node features, which are bag-of-words features from the product descriptions followed by a Principal Component Analysis to reduce the dimension to 100, provided by the OGBN website \cite{OGBNodeProp}.

\subsubsection{OGBN-Protein}\label{datapreprocessingProtein}
The protein samples from OGBN-Protein datasets originate from eight distinct species. In order to appropriately categorize these species within the dataset, we have elected to employ a one-hot encoding strategy. This approach facilitates the conversion of the categorical species data into a binary vector format that can be more readily processed by machine learning algorithms. Each species is represented by a unique binary vector in this schema, where the bit corresponding to a particular species is set to one, while the remaining bits are set to zero.

Fig. \ref{catVec_multigraph} illustrates how we concatenate graph coordinates (GC)%
to get a representation vector for each node in the OGBN-Protein graph. We consider the OGBN-Proteins dataset as a multilayer graph and split eight dimensions into eight graphs $G_i$, $1\leq i \leq 8$, where each graph corresponds to a feature of the edge. We consider the value of 1 minus the confidence of a single association as the weighted distance between two adjacent nodes. Let $d_{jk}$ be the lowest weighted distance between node $j$ and node $k$, and we construct matrix $D_i$ for each graph $i$.

For each graph $G_i$, we follow the method described in Section \ref{Graph Coordinates Introduction} to extract $n_c^i$ topology coordinates (TC) or directional virtual coordinates (DVC).

When extracting TC, while $n_c^i$, the number of coordinates for graph $i$, may be selected based on different criteria, in our experiments presented here, we consider two cases, first is to set the number of coordinates for graph $i$ $n_c^i$ to a constant value over all the graphs, and the second is to set $n_c^i$ 
based on how much variance the topology coordinates capture as described in Section \ref{Graph Coordinates Introduction}. 
For the case of DVC, we use a constant number for $n_c^i$ over all the graphs.
Finally, we concatenate the graph coordinates from 8 graphs with the one-hot encoding as the representation vector for one node, which corresponds to the feature vector in Fig. \ref{catVec_multigraph}. We use the concatenated representation vectors as the inputs for our (non-graph) neural network model. All these steps,  including SVD and concatenation, can be computed very efficiently.

\subsection{Training Configuration}

\subsubsection{OGBN-Product}
We use the train, validation, and test splitting provided by Open Graph Benchmark(OGB). There are 196615 samples in the training set. We use accuracy based on the ground truth to evaluate the performance. The same early stop criterion is applied.

We randomly select $M=1000$ nodes from the largest component as anchor nodes, which results in approximately 55800 nodes out of 2449029 nodes having non-unique coordinates in the TC space.

\subsubsection{OGBN-Protein}
We use the train, validation, and test splitting provided by Open Graph Benchmark(OGB). There are 86600 samples in the training set. We use the Area Under The Curve-Receiver Operating Characteristics(ROC-AUC) to evaluate the performance. An early stop criterion is used to terminate the training process, specifically if  ROC-AUC does not improve on the validation set for $n_{es}$ epochs, then the training process is terminated, and the model with the highest ROC-AUC on the validation set is used on the testing set for performance evaluation. Because the training set has a large number of samples, we also try evaluating the model every $n_{batch}$ mini-batches.

We randomly select $M=1000$ nodes as anchor nodes, which results in approximately 4000 nodes out of 132534 nodes having non-unique coordinates in the TC space. %

\section{Efficacy of Graph Coordinates}\label{Efficacy of Topological Coordinates}
In this section, we demonstrate the efficacy of TCs in capturing the network topology using the two datasets as examples. %
An important question is whether a small set (1-2\%) of randomly selected anchor nodes is able to capture the topology of a network with sufficient detail and robustness for it to be used with conventional NNs instead of using GNNs. 
We evaluate the approximation in two aspects: the number of ambiguous edges and the number of identical topology coordinates. Ambiguous edges represent relationships between individual nodes that are uncertain or unclear due to insufficient number of anchors \cite{mahindre2022link} used for generating TCs. It is extremely unlikely for the TC or DVC at two nodes to be identical unless their virtual coordinates (VC) are identical, and it is sufficient to find identical VC. In our context, ambiguous edges come from the partial distance matrix. In the partial distance matrix, we consider there is an edge if all absolute element-wise distances are less than 1 for the distance vectors of two nodes. The edges represented in this way but do not currently exist in the original graph are ambiguous edges. Due to the potentially large number of latent edges in networks with numerous nodes, we propose a method for approximating the count of ambiguous edges. This approach allows for a more efficient analysis of uncertain relationships in large-scale graphs while maintaining accuracy. We randomly select $n_{edge}$ potential edges that do not currently exist in the graph, where $n_{edge}$ is the number of currently existing edges in the graph. The number of ambiguous edges in these $n_{edge}$ potential edges is used as the approximation evaluation.

For the OGBN-product dataset, the number of currently existing edges is $n_{edge}=61,859,140$; we randomly pick 1000 anchors and $n_{edge}$ potential edges. The number of ambiguous edges in the $n_{edge}$ potential edges is 35247. It is a tiny percentage (only 0.0570\%). Thus, we consider the approximation acceptable. We also randomly pick nine other anchor sets, and each set includes 1000 anchors. The range of numbers of ambiguous edges in the $n_{edge}$ potential edges for these nine anchor sets is 33942 to 36582.
Even if the anchors are randomly sampled, it does not have a large impact on the approximation.

The second evaluation is the number of identical topology coordinates. For the OGBN-product dataset, 55801 nodes have identical topology coordinates. Compared to the total number of nodes (2,449,029) in the graph, it is only a tiny fraction (2.2785\%). Thus, we consider the approximation acceptable.

\begin{figure}[htbp]
    \centerline{\includegraphics[width=0.5\textwidth]{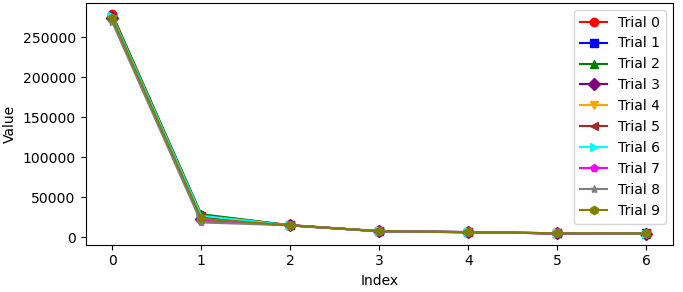}}
    \caption{First 7 singular values for partial distance matrices from 10 randomly sampled anchor sets, each anchor set includes 1000 anchors, dataset: OGBN-Products}
    \label{f7singularValues}
\end{figure}

Even though the anchors are selected at random, they successfully capture the primary characteristics of the entire graph. This is evidenced by the close singular values depicted in Fig. \ref{f7singularValues}. Notably, the variance captured by each coordinate across different anchor sets exhibits slight variation. This property is also observable in the OGBN-protein dataset. 

\section{Experiment Results I - Product Dataset}\label{er2}
The primary purpose of the first set of data is to examine the broad applicability of our method, specifically its capability to function effectively in the context of unweighted graphs.

This first dataset corresponds to an unweighted graph, %
and thus the focus is on the fundamental topology of the graph itself. In this scenario, all edges are considered equal, and the emphasis is on the topological structure rather than the weights of individual connections.

\begin{table}[htbp]
\caption{Number of coordinates, number of trainable parameters and ROC-AUC for each method, dataset: OGBN-Products}
\centering
\begin{tabular}{|c|c|c|c|}
\hline
Method & Coordinates & Parameters & ACC\\
\hline
TCNN10 & 10 & 206,623 & 0.7029 \\
\cline{1-4}
TCNN100A & 100 & 37,039 & 0.7690$\pm$0.0044\\
\cline{1-4}
TCNN100B & 100 & 9,447 & 0.7466$\pm$0.0027\\
\cline{1-4}
DVCNN200/100 &100 & 9,447 & 0.6954$\pm$0.0014 \\
\cline{1-4}

\hline

\end{tabular}

\label{tab:Comparsion_Products}
\end{table}

\subsection{Training with Extracting 10 topology coordinates from the Largest Connected Component}
In the initial stage of our analysis, we extracted a limited subset of topology coordinates with the objective of evaluating the performance of our proposed method. We extract 10 topology coordinates from the graph. We call this experiment case TCNN10. The performance (Accuracy on the testing set) varies between 0.6840 and 0.7029, where the width for hidden layers are [48, 96, 128, 192, 256, 0.1, 256, 128, 64] (0.1 is a dropout layer with drop rate 0.1) for the highest case. In order to get optimal performance, the hidden layers in neural network model need to be complex, thus, the number of parameters is larger as shown in Table \ref{tab:Comparsion_Products}. The training loss curve for the model with the highest accuracy is shown in Fig. \ref{trainCurveCombine} (e). The curve is as smooth as the curve from the OGBN-protein dataset. Even only using a small subset of topology coordinates, our method can still reach acceptable performance.

\subsection{Training with Extracting 100 topology coordinates from the Largest Connected Component}
During the subsequent phase of our investigation, we expanded the quantity of extracted topology coordinates. This strategic augmentation was designed to probe the boundaries of optimal performance, thereby providing a deeper understanding of our method's limitations. 

In our experimental setup, we employed a total of 100 topological coordinates. We call this experiment case TCNN100. The performance, as gauged by the metric of accuracy, fluctuated within the range of 0.7439 (hidden layer: [], we call this configuration TCNN100B) to 0.7734 (hidden layers: [128,64], we call this configuration TCNN100A). The training loss curve for the model with the highest accuracy is shown in Fig. \ref{trainCurveCombine} (f). Notably, even when the number of trainable parameters was constrained to a maximum of 37,039, our model still managed to achieve an impressive average accuracy of 0.7690 (the width for hidden layers are [128,64]). This observation underscores the robustness of our method, highlighting its potential for efficiency even in scenarios of parameter limitation.

As delineated in Section \ref{dataprepro_product}, our previous experiments involved concatenating the extracted topological coordinates with node feature vectors. Interestingly, even when the node feature vectors were excluded from the process, our method remained effective, yielding an accuracy within the range of A to B. This observation reaffirms our hypothesis, demonstrating that our method is capable of capturing the majority of the graph's information. This attribute lends further credence to the robustness and efficacy of our approach, irrespective of the presence or absence of node feature vectors.

\subsection{Training with Extracting 100 DVCs from the Largest Connected Component}
In our experimental evaluation, we implement an experimental setting labeled as DVNN200/100, where we randomly select 200 anchor nodes and form 100 DVCs by pairing every two anchors. Despite a slightly diminished performance (average ACC is only 0.6954) compared to other configurations, as shown by ROC-AUC metric in TABLE \ref{tab:ComparsionLeaderBoard_Products}, this approach yields significant computational benefits.

Primarily, the computational expense is dramatically reduced as we are only required to sample the distance between the chosen 200 anchor nodes and the remaining nodes, as opposed to the 1000 anchors required by our TCS method. Moreover, the DVNN200/100 strategy eliminates the need for SVD, leading to further savings in computational resources. Consequently, this represents a valuable trade-off between performance and computational efficiency, making it a viable option for scenarios where computational resources are limited.

\begin{table}[ht]
\caption{Number of coordinates, number of trainable parameters and ROC-AUC for each method, dataset: OGBN-Proteins}
\centering
\begin{tabular}{|c|c|c|c|}
\hline
Method & Coordinates & Parameters & ROC-AUC\\
\hline
TCNN5 & 40 & 360,400 & 0.6656 \\
\cline{1-4}
TCNN10 & 80 & 364,240 & 0.7950\\
\cline{1-4}
TCNN0.99 &84 & 10,416 & 0.7827$\pm$0.0012 \\
\cline{1-4}
TCNN0.999&3134 & 352,016 & 0.8032$\pm$0.0073 \\
\cline{1-4}
DVCNN400/200 &200 & 180,208 & 0.8008$\pm$0.0008 \\
\cline{1-4}

\hline

\end{tabular}

\label{tab:Comparsion_Proteins}
\end{table}

\section{Experiment Results II - Protein Dataset}\label{er1}
Our goal with the analysis of the second dataset is to demonstrate the versatility of our method. It's important to showcase that the method is not confined to handling unweighted graphs but can also be extended to weighted graphs, thereby expanding its utility across a broader range of graph types. This generalization is critical for the applicability and potential impact of our method in various fields where graphs are used, from network analysis to social science.

In analyzing the second dataset, we maintained the same level of rigor and methodological consistency as the first. We applied our method to the weighted graph, carefully observing how the edge weights influence the method's performance and the resulting topology coordinates.

\subsection{Training with Extracting $n_c^i$ = 5 topology coordinates from Each Graph}
We first experiment on extracting constant 5 topology coordinates from each graph, 40 topology coordinates are extracted in total. After concatenating the extracted topology coordinates with the one-hot encoding, each sample has a 48-dimension representation vector. 

The highest ROC-AUC on the validation set can be reached when only training less than 10 epochs. Even after varying the configuration of hidden dimensions, the highest ROC-AUC on the testing set can only reach 0.6656 (the width for hidden layers are [96,128,256,512,256,128]). Fig. \ref{trainCurveCombine} (a) shows the training loss curve over epochs for the best configuration of hidden dimensions. Although the loss decreases after 10 epochs, the performance (ROC-AUC) on the validation set does not improve anymore. 

\subsection{Training with Extracting $n_c^i$ = 10 topology coordinates from Each Graph}
We then experiment with increasing the number of topology coordinates from each graph to 10. We call this experiment case TCNN10. Fig. \ref{trainCurveCombine} (b) shows the training loss curve over epochs for the best configuration of hidden dimensions. The highest performance (ROC-AUC) on the testing set reaches 0.7950 (the width for hidden layers are [96,128,256,0.1,512,256,128], where 0.1 is a dropout layer with drop rate 0.1), which is much higher than only extracting 5 topology coordinates. If conditions permit, extracting more coordinates from each graph can improve the performance.

\subsection{Training with Captured Variance $p=0.99$}
We also explore the case of extracting a similar number of topology coordinates but wisely selecting them. In this case, the number of extracted coordinates $n_c^i$ are various for each graph, based on the variance of singular values. When capturing $p=0.99$ variance for each graph, we extract 84 topology coordinates from 8 graphs in total. We call this experiment case TCNN0.99. After concatenating the extracted topology coordinates with the one-hot encoding, each sample has a 92-dimension representation vector. Fig. \ref{trainCurveCombine} (c) shows the training loss curve over epochs for the best configuration of hidden dimensions. The highest performance (ROC-AUC) on the testing set reaches 0.7823(the width for hidden layers are [], in other words, the input layer is directly connected to the output layer), and average ROC-AUC is 0.7827, which are close to extracting 10 topology coordinates. The most exciting thing is even though the model does not have any hidden layers, the performance (ROC-AUC) can still reach a close level. However, the training process needs more epochs to find the optimal when no hidden layer exists. Should training time not be a critical factor, adopting a neural network model without hidden layers may be advantageous, as it reduces the computational time during forwarding and storage requirements for parameters.

\subsection{Training with Captured Variance $p=0.999$}
\begin{figure}[htbp]
    \centerline{\includegraphics[width=0.5\textwidth]{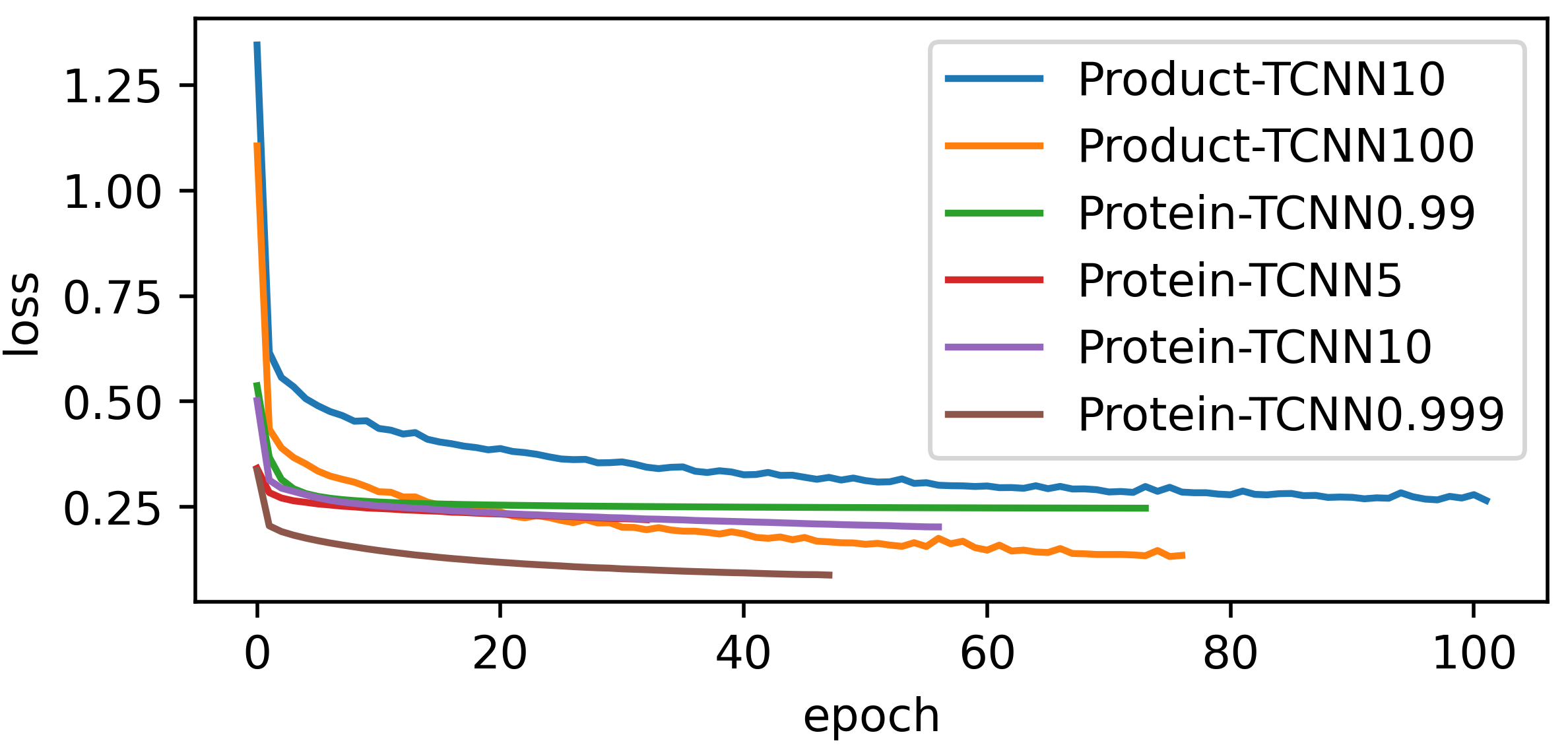}}
    \caption{Training loss curve for (a) Product-TCNN10 (b) Product-TCNN100 (c) Protein-TCNN0.99 (d) Protein-TCNN5 (e) Protein-TCNN10 (f) Protein-TCNN0.999}
    \label{trainCurveCombine}
\end{figure}

However, keeping increasing the number of topology coordinates does not help increase the performance much. When capturing $p=0.999$ variance for each graph, we extract 3134 topology coordinates from 8 graphs in total. We call this experiment case TCNN0.999. After concatenating the extracted topology coordinates with the one-hot encoding, each sample has a 3142 dimension representation vector. \ref{trainCurveCombine} (d) shows the training loss curve over epochs for the best configuration of hidden dimensions. The highest performance (ROC-AUC) on the testing set reaches 0.8103 (the width for hidden layer is [112]), where the average ROC-AUC is 0.8032. The performance only slightly improves compared to capturing 0.99 variance. The performance of the model appears to reach a saturation point, beyond which increasing the number of extracted coordinates yields only marginal improvements. Considering the computational costs associated with enlarging the neural network model, capturing a variance of $p=0.999$ for each graph may be deemed unnecessary. Instead, a more computationally efficient approach might involve capturing a variance of $p=0.99$, which appears sufficient to achieve comparable performance without incurring the additional computational burden.

\subsection{Training with Extracting $n_c^i$ = 200 DVCs from Each Graph}
In our comprehensive experimental evaluation, we also employ an experimental setup called DVNN400/200. This configuration entails randomly selecting the same 400 anchor nodes for all 8 graphs, from which we form 200 DVCs by pairing each of the two anchors. Thus, we have 1,600 DVCs in total. This particular configuration allows us to compare the efficiency between TCS and DVCS.

Comparatively, when juxtaposed with our TCS0.999 method as depicted in TABLE \ref{tab:ComparsionLeaderBoard_Proteins}, we observe that the DVCNN400/200 approach yields an equivalent performance, gauged by ROC-AUC metric (average ROC-AUC is 0.8008). This level of performance is particularly notable given that it is achieved with only 50\% of the trainable parameters required for the TCS0.999 method. Therefore, the DVCNN400/200 configuration illustrates an efficient trade-off between model performance and computational cost, demonstrating the capability of our method to provide accurate results with fewer resources.

Upon thorough analysis of the second dataset, we were able to derive significant insights into the concept that a relatively small number of topology coordinates can efficiently capture the majority of a graph's information. This finding was instrumental in our research as it provided a streamlined approach to understanding the complex structure and properties of graphs. By reducing the complexity of the graph to a manageable number of topology coordinates, we could effectively navigate the graph's intricate patterns and successfully interpret its key aspects.

\begin{table*}[ht]
\caption{Comparsion with the OGBN leaderboard, dataset: OGBN-Products}
\centering
\begin{tabular}{|c|c|c|c|c|}
\hline
Dataset & Method & Rank& ACC & Trainable Parameters \\
\hline
\multirow{7}{*}{Products}&GLEM+GIANT+SAGN+SCR\cite{zhao2022learning}&1  & 0.8737$\pm$0.0006 & 139,792,525\\
\cline{2-5}
&NGraDBERT+GIANT \& SAGN+SLE+CnS\cite{mavromatis2023train} &2 & 0.8692$\pm$0.0007 & 1,154,654\\
\cline{2-5}
& Full-batch GraphSAGE\cite{hamilton2017inductive} & 46 & 0.7850$\pm$0.0014 & 206,895\\
\cline{2-5}
& \textbf{TCNN100A} & & 0.7690$\pm$0.0044 & 37,039\\
\cline{2-5}
&Full-batch GCN\cite{kipf2016semi} &48 & 0.7564$\pm$0.0021 & 103,727\\
\cline{2-5}
&\textbf{TCNN100B} & & 0.7466$\pm$0.0027 & \textbf{9,447}\\
\cline{2-5}
&Node2vec\cite{grover2016node2vec} &51 & 0.7249$\pm$0.0010 & 313,612,207\\
\cline{2-5}
&\textbf{DVCNN200/100} & & 0.6954$\pm$0.0014 & \textbf{9,447}\\
\hline
\end{tabular}

\label{tab:ComparsionLeaderBoard_Products}
\end{table*}

\begin{table*}[ht]
\caption{Comparsion with the OGBN leaderboard, dataset: OGBN-Proteins}
\centering
\begin{tabular}{|c|c|c|c|c|}
\hline
Dataset & Method &Rank& ROC-AUC & Trainable Parameters \\
\hline
\multirow{7}{*}{Proteins}
&GIPA(Wide\&Deep)\cite{li2023gipa} &1 & 0.8917$\pm$0.0007 & 17,438,716\\
\cline{2-5}
&AGDN\cite{sun2020adaptive} &2 & 0.8865$\pm$0.0013 & 8,605,486\\
\cline{2-5}
&GEN+FLAG+node2vec\cite{li2020deepergcn} &16 & 0.8251$\pm$0.0043 & 487,436\\
\cline{2-5}
&\textbf{TCNN0.999} & & 0.8032$\pm$0.0073 & 352,016\\
\cline{2-5}
&\textbf{DVCNN400/200} & & 0.8008$\pm$0.0008 & 180,208\\
\cline{2-5}
& \textbf{TCNN0.99} & & 0.7827$\pm$0.0012 & \textbf{10,416}\\
\cline{2-5}
&GeniePath-BS\cite{liu2020bandit} &17 & 0.7825$\pm$0.0035 & 316,754\\
\cline{2-5}
&GCN\cite{kipf2016semi} &20 & 0.7251$\pm$0.0035 & 96,880\\
\hline

\end{tabular}

\label{tab:ComparsionLeaderBoard_Proteins}
\end{table*}

\section{Comparison with OGBN Leaderboard}\label{section:Comparison}

In our comparative analysis, as indicated in TABLE \ref{tab:ComparsionLeaderBoard_Products} and \ref{tab:ComparsionLeaderBoard_Proteins}, we present a performance benchmark of our models against those on the Open Graph Benchmark Node (OGBN) leaderboard\cite{ogbWebsite}. Specifically, for the OGBN-Proteins dataset, our approach TCS0.99, which is trained by capturing variance with $p=0.99$, utilizes only 10,416 trainable parameters. Despite its relative simplicity, it attains a performance akin to GeniePath-BS, which requires about 30 times the number of trainable parameters. Moreover, when equated with a model using a comparable number of trainable parameters, TCS0.99 exhibits a superior ROC-AUC than the Graph Convolutional Network (GCN)\cite{kipf2016semi}.
Simultaneously, for the OGBN-Products dataset, our TCS100 model, trained by extracting 100 topology coordinates from the largest connected component, necessitates an incredibly minimal number of trainable parameters, i.e., 9,447 (there is no hidden layer). This starkly contrasts with Node2Vec\cite{grover2016node2vec}, which requires 313,612,207 parameters, yet our method achieves a similar accuracy. This comparative study underscores the effectiveness and efficiency of our proposed methods in dealing with graph-structured data.

In conclusion, our study demonstrated that a small subset of randomly selected anchors can effectively capture most of a graph's information. This observation is held across different datasets and edge features, suggesting that the method is generalizable and robust. Our results underscore the graph structure's representational power, allowing even randomly chosen nodes to provide a meaningful representation of the overall graph.

The findings simplify the process of anchor selection, mitigating the risk of overfitting due to overly specific or biased choices and indicating that a graph's informational content is broadly distributed across its nodes. This broad distribution of informational content in the graph structure challenges the notion that complex, carefully curated selection strategies are necessary for effective representation.

\section{Conclusion}\label{conclusion}

A topology coordinate based node embedding scheme combined with conventional neural networks is presented as a computationally efficient alternative for GNNs. We demonstrated the robustness of our approach using benchmark networks corresponding to large weighted and unweighted graphs. Results based on comprehensive experiments on two benchmark datasets, OGBN-Proteins and OGBN-Products confirm the computation efficiency. In particular, our method exhibits a marked performance advantage when considering the number of trainable parameters. Specifically, it attains comparable performance to leading methods on the OGBN leaderboard while utilizing significantly fewer trainable parameters. Conversely, our approach achieves superior performance outcomes given a similar number of trainable parameters. These results highlight our proposed methodology's computational efficiency and practical potential in node property prediction tasks. The effectiveness with which the TCs capture graph topology information is also demonstrated.

The implications of this enhanced capability extends far beyond the immediate sphere of graph-structured data learning. Our method paves the way for a new era of advancements in various fields by making topology utilization more efficient and effective. From the dissection of intricate social networks to the exploration of complex molecular structures in chemistry, the potential applications are vast and varied. It also offers significant potential for breakthroughs in any field where data can be represented as graphs, thereby opening a new frontier in data science.

For future research, these results open up several intriguing avenues. One could explore how this understanding of anchor selection impacts different graph-based algorithms and applications. Another direction could be to investigate how the distribution of information across nodes interacts with different types of graphs and graph properties. Furthermore, examining how this approach scales with larger and more complex graphs could yield insightful results.

Overall, our findings pave the way for more flexible, efficient, and robust methods in graph analysis, potentially leading to transformative impacts across various fields where graph data is prevalent.

\bibliographystyle{IEEEtran}
\bibliography{IEEEabrv,citation}
\vspace{12pt}

\clearpage

\end{document}